%% file: ICLR13.tex
\documentclass{article} 
\usepackage{nips12submit_e,times}

\usepackage{graphicx}
\usepackage{subfigure}
\usepackage[textwidth=2cm,colorinlistoftodos]{todonotes}
\usepackage{xspace}
\usepackage{latexsym}
\usepackage{amsmath,amssymb,graphicx,xspace}
\usepackage{times}   
\usepackage{multirow}
\usepackage{color,comment,rotating,subfigure,url,xspace} 
\usepackage{tikz}
\newcommand\blfootnote[1]{%
  \begingroup
  \renewcommand\thefootnote{}\footnote{#1}%
  \addtocounter{footnote}{-1}%
  \endgroup
}

\usetikzlibrary{arrows,shadows,shapes,backgrounds,decorations,snakes,fit}
\title{Factorized Topic Models}

\author{
\small Cheng Zhang\\
\small CVAP/CAS, KTH\\
\small Stockholm, Sweden \\
\scriptsize\texttt{chengz@kth.se} \\
\And
\small Carl Henrik Ek \\
\small CVAP/CAS, KTH\\
\small Stockholm, Sweden \\
\scriptsize \texttt{chek@kth.se} \\
\And
\small Andreas Damianou\\
\small University of Sheffield\\
\small Sheffield, UK \\
\tiny\texttt{andreas.damianou@sheffield.ac.uk} \\
\And
\small Hedvig Kjellstr\"om\\
\small CVAP/CAS, KTH\\
\small Stockholm, Sweden \\
\scriptsize \texttt{hedvig@kth.se} \\
}

\nipsfinalcopy 
\begin{document}
\maketitle

\vspace{-5mm}
\begin{abstract}
In this paper we present a modif{}ication to a latent topic model, which makes the model
exploit supervision to produce a factorized representation of the
observed data. The structured parameterization separately encodes
variance that is shared between classes from variance that is private
to each class by the introduction of a new prior over the topic space.  
The approach allows for a more eff{}icient inference and provides an
intuitive interpretation of the data in terms of an informative signal
together with structured noise. The factorized representation is shown
to enhance inference performance for image, text, and video
classif{}ication.
\end{abstract}
\blfootnote{This research has been supported by the EU through TOMSY, IST-FP7-
Collaborative Project-270436, and the Swedish Research Council (VR).}

\vspace{-10mm}
\input{introduction}

\input{relatedwork}

\input{model_carl}

\input{experiments}
\input{discussion}

{\footnotesize
\bibliography{ref,carl}}
\bibliographystyle{plain}

\end{document}

%% file: introduction.tex
\section{Introduction}
\label{sec:intro}
\vspace{-2mm}

Representing data in terms of latent variables is an important tool in many applications.  A generative latent variable model provides a parameterization that encodes the variations in the observed data,  relating them to an underlying representation, e.g., a set of classes, using some kind of mapping.  It is important to note that any modeling task is inherently ill-conditioned as there exists  an inf{}inite number of combinations of mappings and parameterizations that could have generated the  data. To that end, we choose different models, based on different assumptions and preferences that will induce different representations, motivated by how well they f{}it the data and for what purpose we wish to use the representation.


Inference in generative models meets diff{}iculties if the variations in the observed data are not representative of the variations in the underlying state to be inferred.  As an example, consider a visual animal classif{}ier, trained with, e.g., SIFT \cite{lowe04} features extracted from training images of horses, cows and cats with a variation of fur texture. The task is now to classify an image of a spotted horse. Based on the features, which will mostly pick up the fur texture, the classif{}ier will be unsure of the class, since there are spotted horses, cows and cats in the training data. The core of the problem is that fur texture is a weak cue to animal class given this data: Horses, cows and cats can all be red, spotted, brown, black and grey. Shape is on the other hand a strong cue to distinguish between these classes. However, the visual features will mostly capture texture information -- the shape information (signal) is ``hidden" among the signif{}icantly richer texture information (structured noise) making up the dominant part of the variation in the data.

In this paper we address this issue by explicitly factorizing the data into a structured noise part, whose variations are shared between all classes, and a signal part, whose variations are characteristic of a certain class. For our purposes, it is very useful to think about data as composed of {\em topics}. Probabilistic topic models \cite{Papadimitriou:2000uq,Hofmann99,blei03,Blei:2012ti} model a data example as a collection of {\em words} (in the case of images, {\em visual words}), each sampled from a latent distribution of {\em topics}. The topics can be thought of as different aspects of the data -- a topic model trained with the data in our animal example above might model one topic for shape and another for fur texture, and a certain data instance is modeled as a combination of a certain shape and a certain texture. 

Our approach is to {\em encourage the topics to assume either a very high correlation or a very low correlation with class}. The class can then be inferred using only the class-specific topics, while the shared topics are used to {\em explain away} the aspects of the data that are not interesting to this particular inference problem.  We present a variant of a Latent Dirichlet Allocation (LDA) \cite{Blei:2012ti} model which is able to  model the signal and structured noise separately from the data. This new model is trained using a factorizing prior, which partitions the topic space into a private signal part and a shared noise part. The model is described in Section \ref{sec:model}.

Experiments in Section \ref{sec:experiments} show that the proposed model outperforms both the standard LDA and a supervised variant, SLDA \cite{blei10}, on classif{}ication of images, text, and video.  Furthermore, the explicit noise model increases the sparsity of the topic representation. This is encouraging for two reasons: f{}irstly, it indicates that the factorized LDA model is a better model of class compared to the unrestricted LDA; enabling better performance on any inference or data synthesis task. Secondly, it enables a more economical data representation in terms of storage and computation; crucial for applications with very large data sets. The factorization method can be applied to other topic models as well, and the sparse factorized topic representation is benef{}icial not only for classif{}ication, as shown here, but also for  synthesis \cite{damianou12}, ambiguity modeling \cite{Ek:2008up}, and domain transfer \cite{navaratnam07}.

%% file: relatedwork.tex
\vspace{-2mm}
\section{Related Work}
\label{relatedwork}
\vspace{-2mm}

In this section we will create a context for the model that we are about to propose  by relating it to factorized latent variable models in general and topic models in specif{}ic. Providing a complete review of either is beyond the scope of this paper, why here we will focus on only the most relevant subset of work needed to motivate the model.

The motivation for learning a latent variable model is to exploit the structure of the new representation to perform tasks such as synthesis or prediction of novel data, or to ease an association task such as classif{}ication. For continuous observations, several classic algorithms such as Principal Component Analysis (PCA) and Canonical Correlation Analysis (CCA) can be interpreted as latent variable models \cite{Bach:2005wz,Lawrence:2005vk,Leen:2006ve,Tipping:1999uo}. Another modeling scenario is when observations are provided in the form of collections of discrete entities. An example is text data where a document consists of a collection of words. One approach to encode such data is using a latent representation that groups words in terms of topics. Several approaches for automatically learning topics from data have been suggested in the literature. A f{}irst proposal of a generative topic model was Probabilistic Latent Semantic Indexing (pLSI) \cite{Hofmann99}. The model represents each document as a mixture of topics. The next important development  in terms of a Bayesian version of pLSI by adding a prior to the mixture weights.  This was done by the adaptation of a Dirichlet layer and referred to as Latent Dirichlet Allocation (LDA) \cite{blei03}.

Central to the work presented in this paper is a specif{}ic latent structure simultaneously proposed by several authors \cite{Ek:2008up,Klami:2006jt,Klami:2008ex,Leen:2008vc}. Given multiple observation modalities of a single underlying state, the purpose of these models is to learn a representation that separately encodes the modality-independent variance from the modality-dependent. The latent representation is factorized such that the modality-independent and modality-dependent are encoded in separate subspaces \cite{Ek:2009vv}. This factorization has an intuitive interpretation in that the private space encodes variations that exists in only one modality and does therefore encode variations representing the ambiguities between the modalities \cite{Ek:2008up}.

In this paper we will exploit a similar type of factorization within a topic model, 
but instead of exploiting correlations between observation modalities, we  employ a single observation modality and a class label associated with each observation.
In specif{}ic, our approach will encourage a factorization relating to class, such that the topics will be split into those encoding within-class variations from those that encode between-class variations. Such  a factorization becomes interesting for inferring the class label from unseen data; the class-shared topics can be considered as representing ``structured noise'' while only the private class topics contain the relevant for class inference.

However, it is not easy to directly transfer the above factorization, formulated \emph{between} modalities and described for continuous data, to topic models, which are inherently discrete.  Results have been presented \cite{jia11,Rasiwasia10,Virtanen12} for the case of two conditionally  independent observation modalities, addressing the image and text cross-modal multimedia retrieval problem with topic representation. In \cite{jia11} a model that can be seen as a Markov random f{}ield of LDA topic models is presented. The topic distribution of each topic model affect the underlying topic spaces of other topic models, connected to that model through the Markov random f{}ield. Further, in \cite{Rasiwasia10}  CCA is applied to the topic space of the text data, which in turn has been learned from LDA and the image feature space. LDA and CCA are used as two separate steps. Differently, \cite{Virtanen12} instead use a Hierarchical Dirichlet Process (HDP) based method which has a complexity selection property. It takes the topics that only describe variance in only one modality as the private space, which explains away the information that cannot be matched between different modalities. This is an extension of \cite{paisley11} to multi-modalities, hence it can not be generalized to other topic models, such as LDA or pLSI and it can not be used to model the private and shared information with only one modality.

Differently from \cite{jia11,Virtanen12,Rasiwasia10}, which need to model the shared topics and private topics in the joint topic space across different observation modalities, our factorization  takes place over one modality across different classes, where the structured noise is modeled in the class-shared topics and the signal is modeled in class-private topics. Furthermore, and importantly, our approach is flexible and can be easily transferred to any type of topic model. Our choice of LDA stems from the fact that it has previously been successfully applied for a large range of data and has desirable sparsity properties that makes for an eff{}icient model.

Topic models, and the LDA model in specif{}ic, are motivated by the benef{}it of representations that are sparse in terms of the distribution of topics for each document. In addition to this, the model we are about to present aims to encourage a specif{}ic structure of the topic themselves. This notion is not new and have been proposed by several other authors. In \cite{gormley12} the topics are represented as combinations of a small number of latent components as such leading to a more compact model. In \cite{wang09} the each topic is constrained by the words in the vocabulary. However, none of these models aim to learn a topic structure that is related to class.

%% file: model_carl.tex
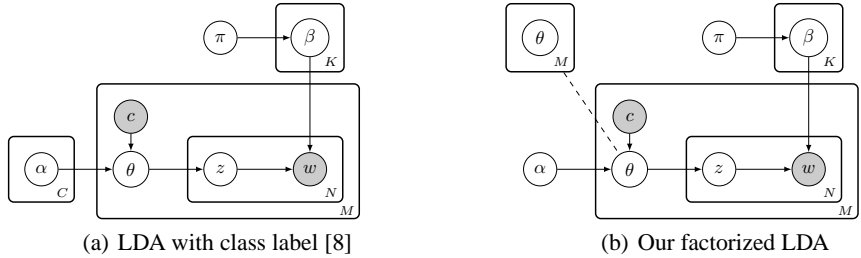
\begin{figure}
\centering
\subfigure[LDA with class label  \cite{feifei05}]{%
\hspace{-10mm}\scalebox{0.7}{\input{CLDA.tex}}} \hspace{25mm}
\subfigure[Our factorized LDA]{%
\hspace{-10mm}\scalebox{0.7}{\input{FLDA.tex}}}
\vspace{-3mm}
\caption{\small Graphic representation of LDA structures. The notation in (b) is adopted from Jia et al.~\cite{jia11}.} 
\label{fig:LDA}
\vspace{-2mm}
\end{figure}

\vspace{-2mm}
\section{Model}
\label{sec:model}
\vspace{-2mm}
As described in the introduction, we add factorization to a model that describes variations of data in terms of a set of latent topics. We seek a structured representation that encodes topics containing within-class, or class private, variations separately from those containing variations that are shared between the classes. We apply our factorization framework to an adaptation of LDA, which incorporates additional class information to recover such a factorized latent space. In this section, the traditional LDA model  \cite{blei03} is f{}irst revisited, followed by the description of our factorized topic model. 

\vspace{-2mm}
\subsection{LDA Revisited}
\vspace{-2mm}
Formally a document $\mathbf{w}$ consist of a collection of words $\mathbf{w}=[w_1,\ldots,w_N]$ from a vocabulary indexed by $\{1,\ldots,V\}$. Within a topic model each document of $N$ words is described as a mixture of $K$ topics such that each word is associated with a specif{}ic topic: $\mathbf{z} = [z_1,\ldots,z_N]$, where $z_n \in \{1,\ldots,K\}$. The mixture is def{}ined as \vspace{-2mm}
\begin{equation}
  p(\mathbf{w}|\mathbf{z},\boldsymbol\beta) = \prod_{n=1}^N\sum_{k=1}^Kp(w_n|z_n,\boldsymbol\beta_k)
\end{equation}
where $\boldsymbol\beta_k$ is the distribution over the vocabulary for topic $k$. The novelty, and the reason for the success, of the LDA model is how the topics $\mathbf{z}$ and the topic vocabulary $\boldsymbol\beta$ are constructed within the framework. The underpinning intuition is that the topics should present a compact representation with $K\ll N$, and that the structure of the topics should be sparse such to achieve a robust and interpretable model. Assuming the topics $\mathbf{z}$ to be governed by a multinomial distribution, $\mathbf{z}\sim \mathit{Multi}(\mathbf{z}|\boldsymbol\theta)$, sparsity can be achieved by choosing the parameters $\boldsymbol\theta$ as governed by a Dirichlet distribution, $\boldsymbol\theta \sim \mathit{Dir}(\boldsymbol\theta|\alpha)$. By the same motivation a Dirichlet prior is placed over the topic-vocabulary distribution $\boldsymbol\beta\sim \mathit{Dir}(\boldsymbol\beta|\pi)$. As the Dirichlet is conjugate to the multinomial distribution, the marginal likelihood can be reached analytically by combining the likelihood with the prior and performing the integration, \vspace{-3mm}
\begin{equation}
p(\mathbf{w} | \alpha,\pi) = 
\int p(\boldsymbol\theta | \alpha) \left( \prod_{n=1}^N \left[ \sum_{z_n} p(z_n | \boldsymbol\theta) p(w_n | z_n, \boldsymbol\beta_{z_n}) \right] p(\boldsymbol\beta|\pi) \right) d\boldsymbol\theta
\label{eq:LDA}
\end{equation}
from which the parameters of the model can be learned.

One way of incorporating class information within the LDA framework was suggested in \cite{feifei05} where the use of a class dependent topic distribution was proposed. This was implemented by using the class variable $c$ as a ``switch''; $p(\boldsymbol\theta|\mathbf{\boldsymbol\alpha},c) = \prod_{j=1}^C \mathit{Dir}(\boldsymbol\theta|\alpha_j)^{\delta_{cj}}$ where $\delta_{ij}$ is the Kronecker delta function. Using this model the class can be inferred for a new document $\mathbf{w}^*$ through a maximum likelihood procedure $\hat{c} = \arg\max_c p(\mathbf{w}^* | \boldsymbol\alpha, \pi, c)$ \cite{feifei05}.

In this paper we take inspiration from the work presented in \cite{feifei05}. However, we choose to incorporate the class information in a slightly different manner. In specif{}ic, we use a factorizing prior over the topic distribution, which f{}irstly encourages sparsity, and secondly introduces a preference for a class conditioned structure, such that separate topics encode within-class variations and between-class variations in the data. Thus, the model we will propose have a stronger class dependency compared to \cite{feifei05}. We will now proceed to describe and motivate the relevance of this class dependency.

\vspace{-2mm}
\subsection{Factorized Topic Model}
\vspace{-2mm}
As motivated in Section \ref{sec:intro}, our idea is to separate the topic space into two parts, where the class-private part explains the class-dependent information (signal) and the shared part explains the class-independent information (structured noise). To achieve this we introduce an additional prior $p(\boldsymbol\theta)$ to the model presented in \cite{feifei05}. This will encourage a factorized structure such that the $K$ topics can be ``softly'' split into $K_p$ class-private topics and $K_s$ shared topics where $K_p + K_s = K$. The advantage of such a structured topic space is that it will be more compact than a regular model; all aspects of the data that correlate with class will be pushed into the class-private part of the topic space. Since the other, class-shared, part of the topic space will then only contain noise, the class of a new document $\mathbf{w^*}$ will in effect be inferred using only the class-private part. Further, in our model, we will use the same sparsity prior $\alpha$ over the topics for all classes. This removes the additional flexibility of allowing a different topic sparsity for each class -- which can be relevant in certain special cases --- but the gain is a more robust model with fewer free parameters, requiring less training data.

In the following, let $\boldsymbol\theta^\mathrm{class}$ be the topic distributions of all classes, obtained by marginalizing $\theta$ over class. Its rows are def{}ined as $\boldsymbol\theta^\mathrm{class}_c \propto \sum_{m=1}^M {\boldsymbol\theta_m}\delta_{c_mc}$, where $c_m$ indicates the class label of the $m^\mathrm{th}$ document and $\delta_{ij}$ the Kronecker delta function.  Examples of $\boldsymbol\theta^\mathrm{class}$ distributions can be seen in Figures \ref{fig:toy}, \ref{fig:reuters}, \ref{fig:scene}, and \ref{fig:KTH}.

Intuitively, the private topics would concentrate to a certain class in $\boldsymbol\theta^\mathrm{class}$, while the shared topics would be more spread among all classes (more uniformly distributed over a column in $\boldsymbol\theta^\mathrm{class}$). Information entropy, widely used in different f{}ields \cite{leibe08,pluim00}, provides a good measurement of this property. In this case, we employ an entropy-like measure $H(k)$ over class for each topic $k$: \vspace{-1mm}
\begin{equation}
H(k)=\frac{1}{\log(1/C)}\sum^C_{c=1} \frac{\theta^\mathrm{class}_{c,k}}{\sum^C_{\xi=1}\theta^\mathrm{class}_{\xi,k}} \log(\frac{\theta^\mathrm{class}_{c,k}}{\sum^C_{\xi=1}\theta^\mathrm{class}_{\xi,k}}) 
\label{eq:h}
\end{equation}
where $\theta^\mathrm{class}_{c,k}$ is the element in row $c$ and column $k$ of $\boldsymbol\theta^\mathrm{class}$. $H(k) \in [0,1]$, 0 if all the probability in the topic $k$ is concentrated to one class, 1 if all classes are equally probable to contain the topic $k$. 

To split the topics into a private and shared part, we wish the prior $p(\boldsymbol\theta|\kappa)$ to encourage topics $k$ to either have a low $H(k)$ (be very class-specif{}ic) or high $H(k)$ (be very class-unspecif{}ic). Hence, we introduce a function as: 
\begin{equation}
\label{eq:AK1}
A(k)= H(k)^2 - H(k) +1 ~.
 \vspace{-1mm}
 \end{equation}
The prior is def{}ined as: \vspace{-2mm}
\begin{equation}
\label{eq:priordef}
p(\boldsymbol\theta) \propto \prod_{k=1}^K A(k) ~.
\end{equation}
This prior thus treats each column of $\boldsymbol\theta^\mathrm{class}$ independently.
With the additional prior (Figure \ref{fig:LDA}(b)), the generative model becomes: \vspace{-1mm}
\begin{equation}
p(\mathbf{w} | \alpha, \pi, c) = 
\int p(\boldsymbol\theta | \alpha, c) p(\boldsymbol\theta) \left( \prod_{n=1}^N \left[ \sum_{z_n} p(z_n | \boldsymbol\theta) p(w_n | z_n, \boldsymbol\beta) \right] p(\boldsymbol\beta|\pi) \right) d\boldsymbol\theta ~.
 \label{eq:FLDA}
\end{equation}

\vspace{-2mm}
\paragraph{Learning.}
We use Gibbs sampling for learning the parameters of the model, more specifically, collapsed Gibbs sampling \cite{heinrich08} in the same manner as \cite{jia11}.  The factorizing prior presents itself in the learning as an additional factor in the objective function over $z$, compared to the original LDA model. It should be noted that the factorizing prior in Equation \ref{eq:priordef} is independent of the type of learning procedure -- the model in Equation \ref{eq:FLDA} can also be trained using, e.g., a variational method.

When training the model, the topics are initialized randomly, which means that they all have a $H$ close to 1. During Gibbs sampling, it would be very unlikely to f{}ind a topic with low $H$, given the bimodality of $A$ in Equation (\ref{eq:AK1}). To address this problem, we introduce an ``auto-annealing" procedure, where $A$ is replaced with a dynamic cooling function starting off by encouraging low $H$ only, and gradually encouraging high $H$ more and more, as the average $H$ decreases (i.e., when some topics have found a class-specif{}ic state). Hence, $A$ is changed to a dynamic function
\begin{equation}
\label{eq:optEtp}
A(k)= H(k)^2 - 2\hat{H}H(k) +1 
\end{equation}
where the average $H$, $\hat{H}= \sum_{k=1}^{K} H(k) / K$,  is used as an annealing parameter in the function. 
As with other annealing procedures, the ``auto-annealing" procedure means that the factorizing prior $p(\boldsymbol\theta)$ changes in each step of the iterative learning procedure. In a normal annealing procedure, this change would be actuated by changing the annealing parameter. Here, $\hat{H}$ can be thought of as an autonomous annealing parameter since it converges automatically to a value reflecting the fraction of the class-dependent versus class-independent variation in the data. For example, the text data set (Figure \ref{fig:reuters}) has a lower $\hat{H}$ than the natural scene dataset (Figure \ref{fig:scene}).

\vspace{-2mm}
\paragraph{Segmenting the topic space.}
When the model have been trained we can evaluate the structure of the learned topic space by computing $H(k)$ for each topic $k$. We consider topics with low $H$ as class-dependent while topics with high $H$ are considered as independent. As such the topic space can be ``softly'' segmented and interpreted in a class conditioned manner. As an example, the words building up the shared topics can be considered as {\em stop words}. In text processing, there is usually some standard stop words list, which can be used to pre-process the text. However, these stop words are predef{}ined, for example, ``the", ``at" etc. However, they sometimes also provide class-relevant information, for example, some topics are more location dependent or have more nouns. On the other hand, there are words, like ``learning'', ``performance'' etc, which do not carry much information in, say, a machine learning conference corpus.  In our model, we automatically learn the real stop words for the given domain. Furthermore, while it is easy to predef{}ine the stop words  in text data, this problem becomes much more challenging in computer vision applications. The  ``stop-visual-words'' are ill-def{}ined and much less intuitive to f{}ind, why an algorithm which automatically learns them, such as the one we propose, is very benef{}icial. We would like to emphasize again that there is still only one topic space; no hard splitting or removal of topics is done, neither for learning, nor for inference.

%% file: CLDA.tex
\pgfdeclarelayer{background}
\pgfdeclarelayer{foreground}
\pgfsetlayers{background,main,foreground}

\begin{tikzpicture}

\tikzstyle{surround} = [thick,draw=black,rounded corners=1mm]

\tikzstyle{scalarnode} = [circle, draw, fill=white!11,  
    text width=1.2em, text badly centered, inner sep=2pt]

\tikzstyle{discnode}=[rectangle,draw,fill=white!11,minimum size=0.9cm]

\tikzstyle{vectornode} = [circle, draw, fill=white!11,  
    text width=2.3em, text badly centered, inner sep=2pt]
\tikzstyle{state} = [rectangle, draw, text centered, fill=white, 
    text width=8em, text height=6.7em, rounded corners]

\tikzstyle{arrowline} = [draw,color=black, -latex]
\tikzstyle{carrowline} = [line width=2pt, draw,color=black, -latex]
\tikzstyle{line} = [draw]

\node [scalarnode] at (0,0) (alpha) {$\alpha$};
\node [scalarnode] at (1.7, 0) (theta) {$\theta$};
\node [scalarnode, fill=gray!40] at (1.7, 1) (c) {$c$};
\node [scalarnode] at (3.4, 0) (z) {$z$};
\node [scalarnode, fill=gray!40] at (5.1, 0) (w) {$w$};
\node [scalarnode] at (5.1, 2.5) (beta) {$\beta$};
\node [scalarnode] at (3.4, 2.5) (pi) {$\pi$};

\node[surround, inner sep = .3cm] (f_alpha) [fit = (alpha)] {};
\node[surround, inner sep = .3cm] (f_zw) [fit = (z) (w)] {};
\node[surround, inner sep = .3cm] (f_eta) [fit = (c) (theta) (f_zw) ] {};
\node[surround, inner sep = .28cm] (f_beta) [fit = (beta)] {};

\node [] at (0+0.4,0-0.45) (C) {\scriptsize $C$};
\node [] at (5.1+0.4,0-0.45) (N) {\scriptsize $N$};
\node [] at (5.1+0.7,0-0.78) (M) {\scriptsize $M$};
\node [] at (5.1+0.4,2.5-0.45) (K) {\scriptsize $K$};

\path [arrowline] (c) to (theta); 
\path [arrowline] (alpha) to (theta);  
\path [arrowline] (theta) to (z);
\path [arrowline] (z) to (w);
\path [arrowline] (beta) to (w);
\path [arrowline] (pi) to (beta);

\end{tikzpicture}


%% file: FLDA.tex
\pgfdeclarelayer{background}
\pgfdeclarelayer{foreground}
\pgfsetlayers{background,main,foreground}

\begin{tikzpicture}

\tikzstyle{surround} = [thick,draw=black,rounded corners=1mm]

\tikzstyle{scalarnode} = [circle, draw, fill=white!11,  
    text width=1.2em, text badly centered, inner sep=2pt]

\tikzstyle{discnode}=[rectangle,draw,fill=white!11,minimum size=0.9cm]
\tikzstyle{vectornode} = [circle, draw, fill=white!11,  
    text width=2.3em, text badly centered, inner sep=2pt]
\tikzstyle{state} = [rectangle, draw, text centered, fill=white, 
    text width=8em, text height=6.7em, rounded corners]

\tikzstyle{arrowline} = [draw,color=black, -latex]
\tikzstyle{carrowline} = [line width=2pt, draw,color=black, -latex]
\tikzstyle{line} = [draw]

\node [scalarnode] at (0,0) (alpha) {$\alpha$};
\node [scalarnode] at (1.7, 0) (theta) {$\theta$};
\node [scalarnode, fill=gray!40] at (1.7, 1) (c) {$c$};
\node [scalarnode] at (3.4, 0) (z) {$z$};
\node [scalarnode, fill=gray!40] at (5.1, 0) (w) {$w$};
\node [scalarnode] at (5.1, 2.5) (beta) {$\beta$};
\node [scalarnode] at (3.4, 2.5) (pi) {$\pi$};
\node [scalarnode] at (0, 2.5) (theta2) {$\theta$};

\node[surround, inner sep = .3cm] (f_zw) [fit = (z) (w)] {};
\node[surround, inner sep = .3cm] (f_eta) [fit = (c) (theta) (f_zw)] {};
\node[surround, inner sep = .28cm] (f_beta) [fit = (beta)] {};
\node[surround, inner sep = .3cm] (f_theta2) [fit = (theta2)] {};

\node [] at (5.1+0.4,0-0.45) (N) {\scriptsize $N$};
\node [] at (5.1+0.7,0-0.78) (M) {\scriptsize $M$};
\node [] at (5.1+0.4,2.5-0.45) (K) {\scriptsize $K$};
\node [] at (0+0.4,2.5-0.45) (M2) {\scriptsize $M$};

\path [arrowline] (c) to (theta); 
\path [arrowline] (alpha) to (theta);  
\path [arrowline] (theta) to (z);
\path [arrowline] (z) to (w);
\path [arrowline] (beta) to (w);
\path [line,dashed] (f_theta2) to (theta);
\path [arrowline] (pi) to (beta);

\end{tikzpicture}


%% file: experiments.tex
\vspace{-2mm}
\section{Experiments}
\label{sec:experiments}
\vspace{-2mm}

\begin{figure}[!t]
\begin{center}
\includegraphics[width=0.97\columnwidth]{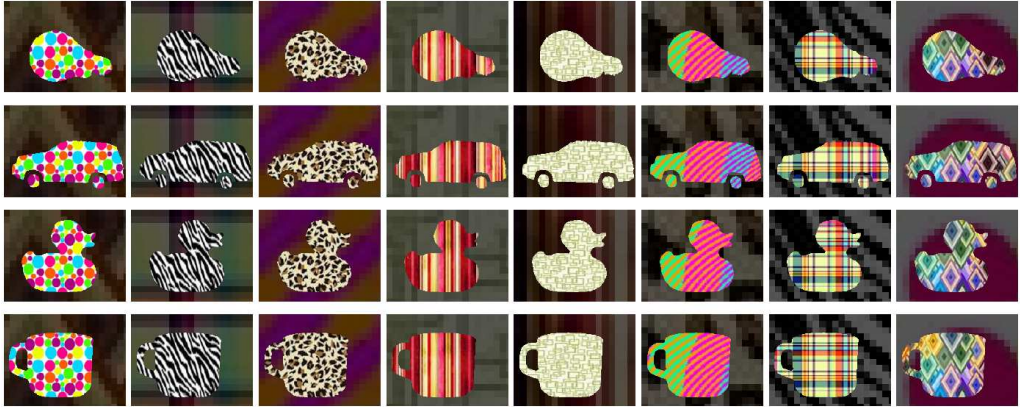}
\vspace{-2mm}
\caption{\small All the instances in the toy object dataset. }
\label{fig:ToyImgs}
\end{center}
\vspace{-3mm}
\end{figure}

The proposed model is evaluated on four different classif{}ication tasks, and
compared to two baselines consisting of a regular LDA model with class label
\cite{feifei05}, and a model with stronger class-supervision in the topic learning, SLDA \cite{blei10}.

\begin{figure}[t]
\centerline{%
\subfigure[\small $\boldsymbol\theta^\mathrm{class}$ with regular LDA]{%
\includegraphics[width=0.33\columnwidth]{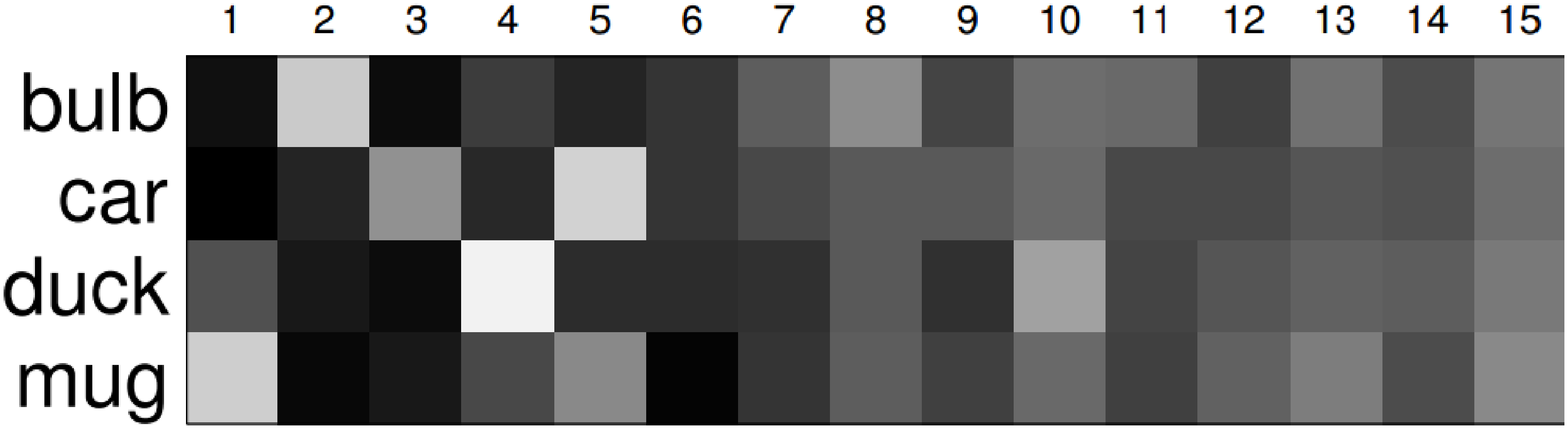}
\label{fig:etp0}}
\subfigure[\small $\boldsymbol\theta^\mathrm{class}$ with SLDA]{%
\includegraphics[width=0.33\columnwidth]{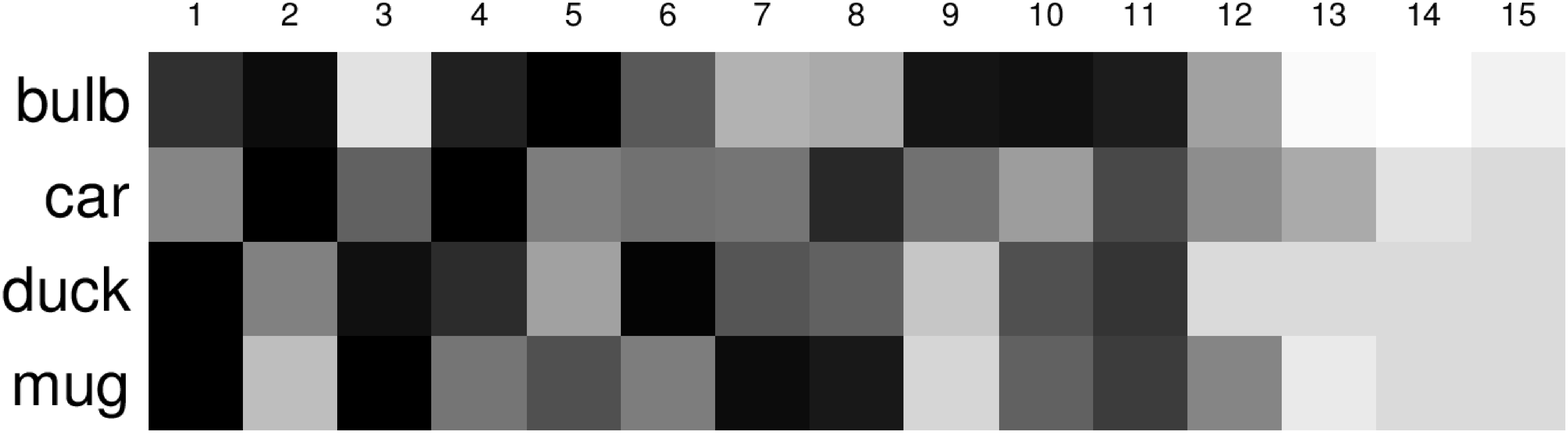}
\label{fig:etpslda}}
\subfigure[\small $\boldsymbol\theta^\mathrm{class}$ with factorized LDA]{%
\includegraphics[width=0.33\columnwidth]{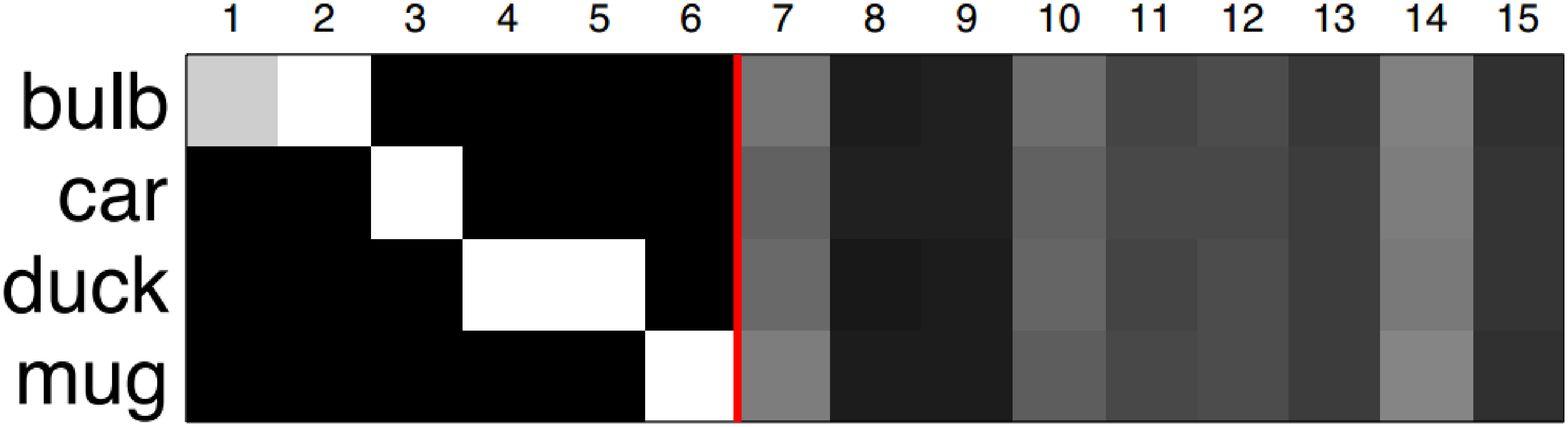}
\label{fig:etp1}}}
\vspace{-3mm}
\caption{\small Toy object dataset. (a) Regular LDA topic distribution
marginalized over class $\boldsymbol\theta^\mathrm{class}$, topics sorted in
ascending order of class-specif{}icity. (b) SLDA topic distribution
marginalized over class $\boldsymbol\theta^\mathrm{class}$, topics sorted in
ascending order of class-specif{}icity. (c) Factorized LDA topic distribution
marginalized over class $\boldsymbol\theta^\mathrm{class}$, topics sorted in
ascending order of class-specif{}icity, red line indicating partition between
$\boldsymbol\theta^p$ and $\boldsymbol\theta^s$. }
\label{fig:toy}
\vspace{-2mm}
\end{figure}

\vspace{-2mm}
\subsection{Object Classif{}ication}
\vspace{-2mm}
We f{}irst demonstrate how the factorization works using a toy dataset. The
dataset, shown in Figure \ref{fig:ToyImgs}, is constructed to have a very high
degree of structured noise. There are four object classes: bulb, car, duck, and
mug. All 8 instances of a certain class have the same shape and image location.
However, there is a very high intra-class variability in foreground and
background texture. Furthermore, all four classes contain the exact same
foreground-background texture combinations. Thus, the texture (which will
dominate the variation among features from any visual extractor) can be regarded
as structured noise, while the true signal relates to shape. 
The properties of this dataset can also be found to some extent in natural
images: most realistic image and object classes display large intra-class
appearance variation, and different classes share appearance aspects.
Furthermore, the backgrounds in natural scenes are often complex and varying,
introducing even more variation among training data for a class.

SIFT features on two different scales are densely extracted from all images, and
a 64-word vocabulary is learned in which all SIFT features are represented.
Thus, each image is represented by a bag of visual words in this vocabulary.

The experiment is performed in a  hold-one-out manner, where each image in turn
is classif{}ied using a model trained on the other 31 images. In the following,
we
will by ``regular LDA" mean the regular LDA with upstream supervision presented
in \cite{feifei05}, but trained using Gibbs sampling in the same way as our
model, with the same value of $\alpha$ for all documents. With "SLDA", we mean the more 
strongly supervised LDA variant with downstream supervision presented in \cite{blei10}, 
implemented by Blei et al.

Our proposed factorized LDA, as well as regular LDA and SLDA, are trained with 15 topics,
$\alpha=0.1$ and $\pi=0.2$. 
The classif{}ication performance for each class is found by averaging over the
performances for the 8 images of that class. 

It should be noted that the test image always will have a texture that is
different from the training images of that class. However, the same texture can
be found in other classes. A classif{}ier that tries to explain all variation in
the data in terms of class variation will therefore have diff{}iculties in
modeling this data set; a regular LDA or SLDA model trained with
this data will be forced to represent texture as well as shape in the same topics,
since the Dirichlet prior will promote topic sparsity. Thus, very few topics
will purely represent one class, as shown in Figures \ref{fig:etp0} and \ref{fig:etpslda}.

However, our model, which explicitly factorizes the topics into those private to
a certain class and those shared between all classes, will allow the relevant
shape variation to be represented separately from the texture variation, which
will just confuse the classif{}ication in this case. Figure  \ref{fig:etp1}
shows
the factorized topic distribution; it is clear that the topics in
$\boldsymbol\theta^p$ are private to a certain class, while the noise topics in
$\boldsymbol\theta^s$ are shared equally over all classes; all the structured
noise has thus been pushed into $\boldsymbol\theta^s$. Thus, even though the
full topic space is used for classif{}ication, it is effectively only based on
$\boldsymbol\theta^p$, while the shared topics $\boldsymbol\theta^s$ (right of
the red line in Figure \ref{fig:etp1}) are effectively disregarded in the
classif{}ication since they appear with equal probability in all classes.

As expected, the explicit noise model greatly improves classif{}ication on this
dataset: the factorized LDA
reaches 81.25\%,  while a regular LDA reaches
a classif{}ication rate of 34.38\%, only slightly above chance, and SLDA who is forced by the stronger supervision to represent all variation (where texture is dominating) in terms of class achieves a result of 0\% since the texture of the test image is not present in the training data of the same class.

\begin{figure}[!t]
\centerline{%
\subfigure[\small $\boldsymbol\theta^\mathrm{class}$ with regular LDA]{
\includegraphics[width=0.32\columnwidth]{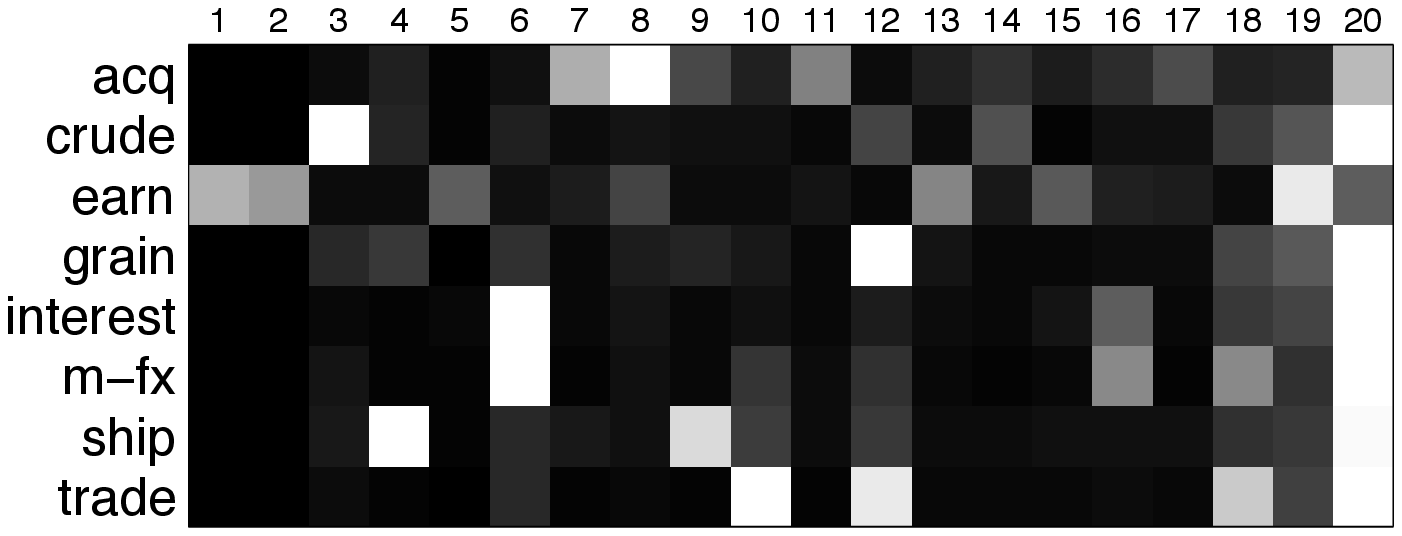}
\label{fig:R8etp0}}
\subfigure[\small $\boldsymbol\theta^\mathrm{class}$ with SLDA]{
\includegraphics[width=0.32\columnwidth]{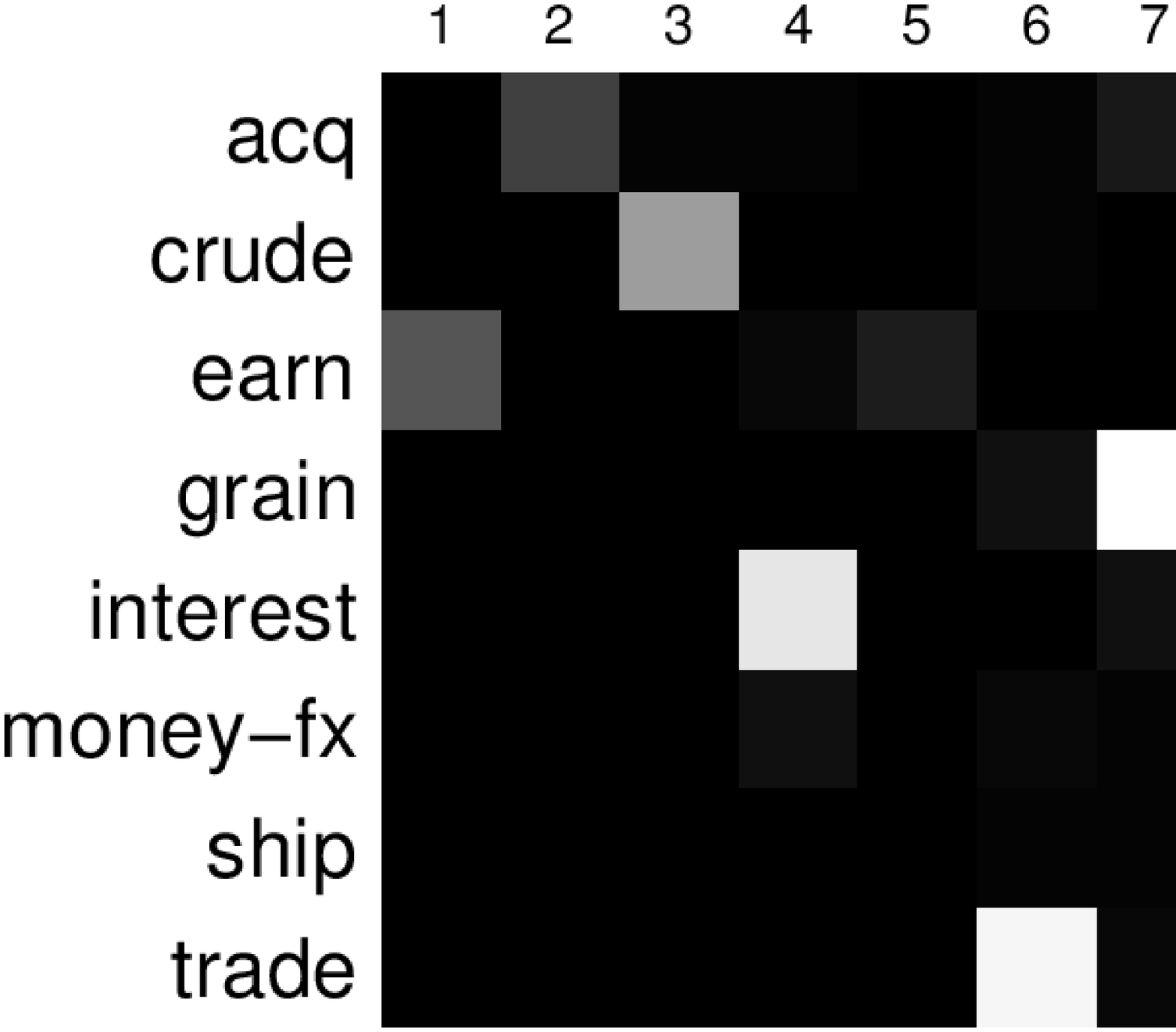}
\label{fig:R8slda}} 
\subfigure[\small $\boldsymbol\theta^\mathrm{class}$ with factorized LDA]{
\includegraphics[width=0.32\columnwidth]{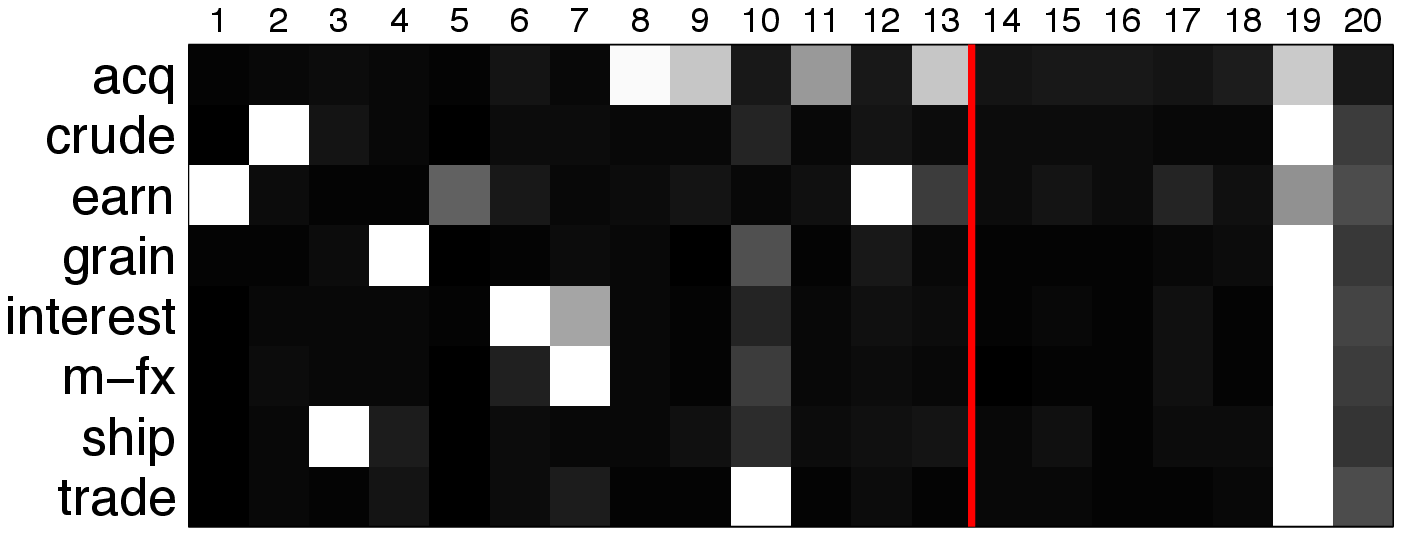}
\label{fig:R8etp1}}}
\vspace{-3mm}
\caption{ \small Reuters 21578 R8 dataset. (a) Regular LDA topic distribution
marginalized over class $\boldsymbol\theta^\mathrm{class}$, topics sorted in
ascending order of class-specif{}icity. (b) SLDA topic distribution
marginalized over class $\boldsymbol\theta^\mathrm{class}$, topics sorted in
ascending order of class-specif{}icity. (c) Factorized LDA topic distribution
marginalized over class $\boldsymbol\theta^\mathrm{class}$, topics sorted in
ascending order of class-specif{}icity, red line indicating partition between
$\boldsymbol\theta^p$ and $\boldsymbol\theta^s$. }
\label{fig:reuters}
\end{figure}

\vspace{-2mm}
\subsection{Text Classif{}ication}
\vspace{-2mm}
We now evaluate the proposed model in a realistic text classif{}ication
scenario.
We use the standard R8 training and testing set from the Reuters 21578 dataset
\cite{reuters}, which contains 5485 training documents and 2189 testing
documents. The all-terms version of the data is used since we want to illustrate
how our model deals with noise.

The regular LDA, SLDA and factorized LDA models are trained with 20 topics, and
parameter settings $\alpha=0.5$ and $\pi=0.1$.
The topic distributions are shown in Figures \ref{fig:R8etp0},  \ref{fig:R8slda}, and \ref{fig:R8etp1}. The
factorized class-private topic distribution $\boldsymbol\theta^p$ (left of the
red line in Figure \ref{fig:R8etp1}) is noticeably cleaner than the regular
distribution $\boldsymbol\theta^\mathrm{class}$ (Figure \ref{fig:R8etp0}). In
the factorized LDA, only $\boldsymbol\theta^p$ contributes to the
classif{}ication, while the shared topics $\boldsymbol\theta^s$ (right of the
red
line in Figure \ref{fig:R8etp1}) are effectively disregarded since they appear
with equal probability in all classes. The topics of the SLDA model are sparser (Figure \ref{fig:R8slda}), but all topics are forced to be class-specif{}ic by the stronger supervision.

There is a signif{}icant classif{}ication improvement using the factorized topic
space, from 74.63\% with regular LDA  and 63.75\% with SLDA
to 83.91\% with factorized LDA.

\begin{figure}[t]
\centerline{\subfigure[\small $\boldsymbol\theta^\mathrm{class}$ with regular
LDA]{\includegraphics[width=0.33\columnwidth]{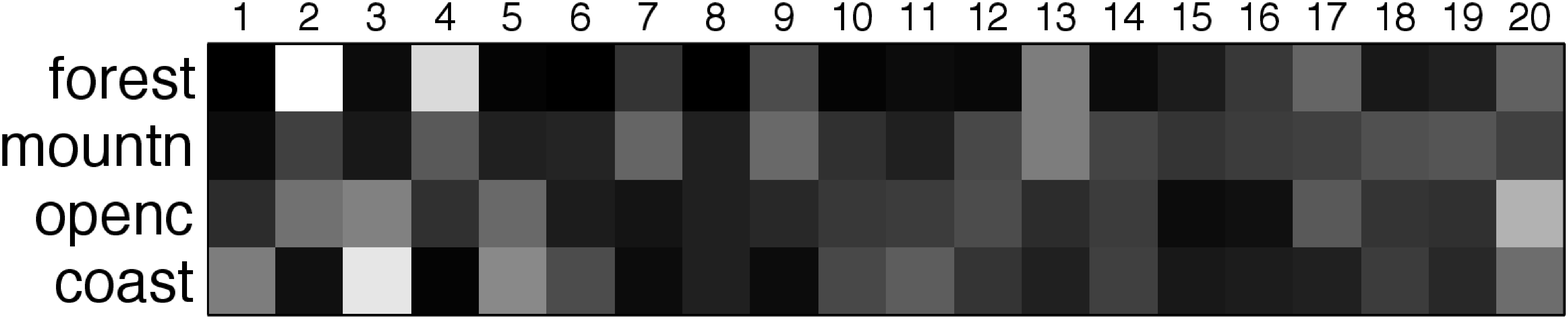}
\label{fig:Setp0}}
\subfigure[\small $\boldsymbol\theta^\mathrm{class}$ with SLDA]{
\includegraphics[width=0.33\columnwidth]{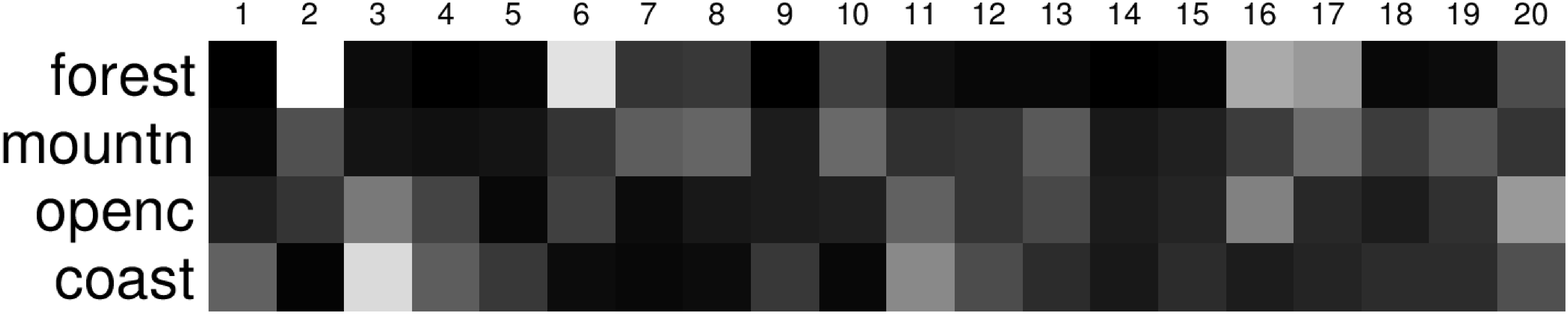}
\label{fig:SetpSLDA}}
\subfigure[\small $\boldsymbol\theta^\mathrm{class}$ with factorized LDA]{
\includegraphics[width=0.33\columnwidth]{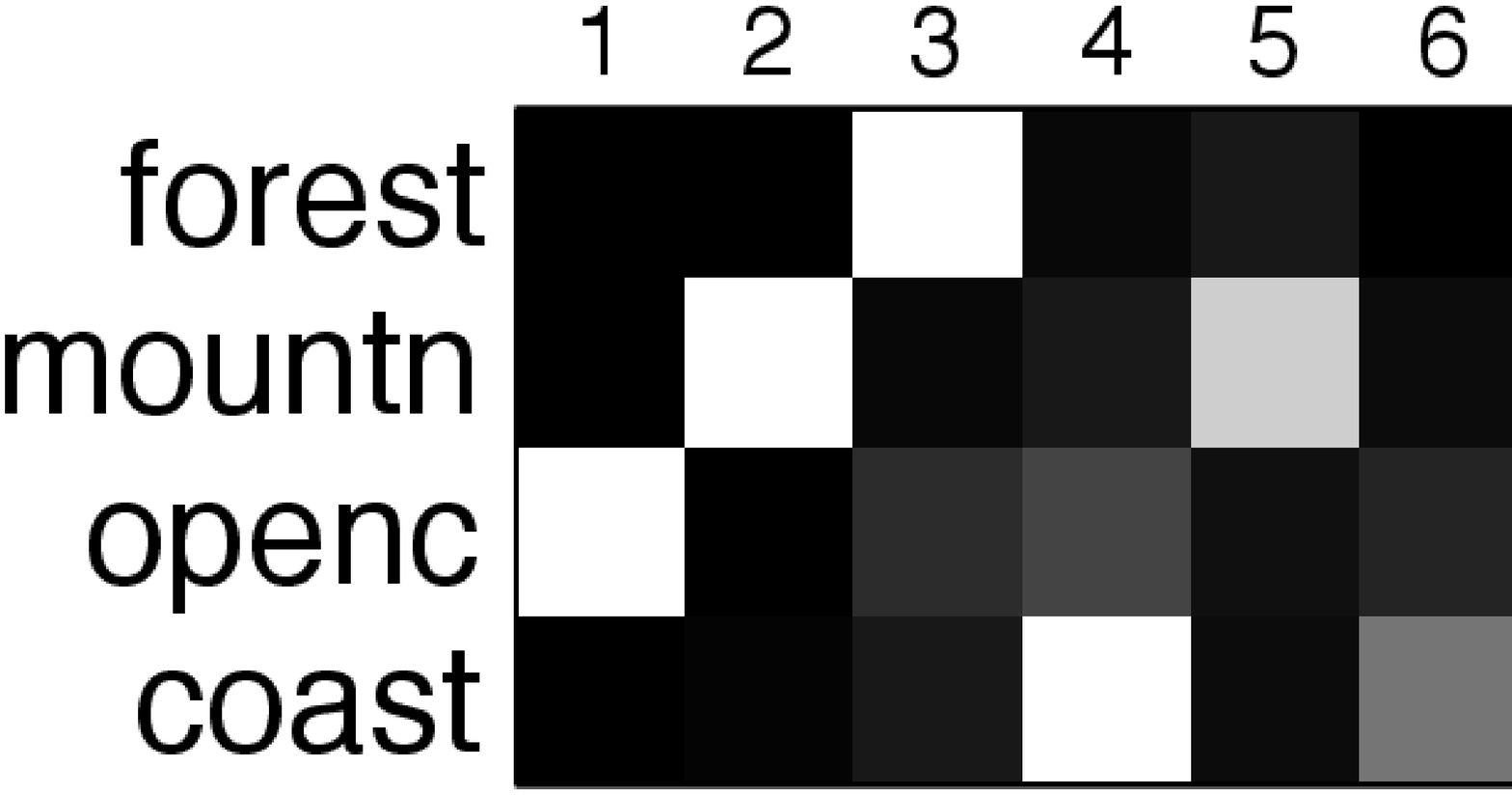}
\label{fig:Setp1}}}
\vspace{-3mm}
\caption{ \small Natural scene dataset. (a) Regular LDA topic distribution
marginalized over class $\boldsymbol\theta^\mathrm{class}$, topics sorted in
ascending order of class-specif{}icity. (b) SLDA topic distribution
marginalized over class $\boldsymbol\theta^\mathrm{class}$, topics sorted in
ascending order of class-specif{}icity. (c) Factorized LDA topic distribution
marginalized over class $\boldsymbol\theta^\mathrm{class}$, topics sorted in
ascending order of class-specif{}icity, red line indicating partition between
$\boldsymbol\theta^p$ and $\boldsymbol\theta^s$.}
\label{fig:scene}
\vspace{-2mm}
\end{figure}

\vspace{-2mm}
\subsection{Scene Classif{}ication}
\vspace{-2mm}
We also evaluate the proposed model on a challenging natural scene dataset used
in \cite{feifei05}. There are four classes: forest, mountain, open country and
coast, with 100 training images and 50 test images per class. From each image,
SIFT features on two different scales are densely extracted, and labeled
according to a 192-word vocabulary learned from the features, as in
\cite{feifei05}.

The regular LDA, SLDA, and factorized LDA models are trained with 20 topics, and
parameter settings $\alpha=0.5$ and $\pi=0.1$.
Figures \ref{fig:Setp0}, \ref{fig:SetpSLDA}, and \ref{fig:Setp1} show the respective topic distributions;
notably, the class-specif{}ic topic space $\boldsymbol\theta^p$ effectively used
for classif{}ication in our factorized LDA only contains 8 topics, while 12
topics
($\boldsymbol\theta^s$) are devoted to modeling structured noise. Thus, the
factorized representation is notably sparser than a regular LDA representation,
which gives the opportunity to save both storage space and computation time
during classif{}ication -- an important factor to take into account for large
datasets.

In addition to rendering a notably sparser data representation, the factorized LDA reaches a marginally higher performance rate than with a regular LDA and SLDA: 84.50\% for our model compared to 80.50\% for the regular LDA and 84.00\% for SLDA. All performances are slightly better than the original implementation of the regular LDA \cite{feifei05}, which reaches 76.0\%.

\vspace{-2mm}
\subsection{Action Classif{}ication}
\vspace{-2mm}

We proceed to evaluate the methods on a dataset with more variation independent of class. The dataset consists of three actions from the KTH Action dataset \cite{laptev05}: boxing, handclapping and handwaving. There are 100 short video sequences of each action, which show 25 different people performing the action, recorded in four shooting conditions (zooming and panning of camera, different background ). The shooting condition has large influence on the motion in the video, as each zoom or panning motion adds global motion to the video and backgrounds contribute to the motion features as well. However, the variation in shooting condition is not at all correlated with action class in the dataset. Just as in the toy experiment above (but now in a more realistic setting), a large proportion of the data variation is thus independent of the action class. Due to the low signal-to-noise ratio, a topic model without factorization will have diff{}iculties capturing the aspects of data relevant for discriminating activity class.

The experiment was performed by separating out from the training data all the 25 images of an action f{}ilmed with a certain shooting condition. The topic models were then trained with all other data, and evaluated with the 25 removed images. Hence, the certain combination of action and camera condition in the test data was not present in the training data.
This was done for all actions  in turn, and the result was averaged over actions.

STIP features \cite{laptev05} were extracted from all sequences and clustered into a vocabulary of 128 spatio-temporal words. This representation was used to train the regular LDA, SLDA and factorized LDA models with 10 topics, $\alpha = 0.1$ and $\pi = 0.1$.

\begin{figure}[t]
\centerline{\subfigure[\small $\boldsymbol\theta^\mathrm{class}$ with regular
LDA]{
\includegraphics[width=0.33\columnwidth]{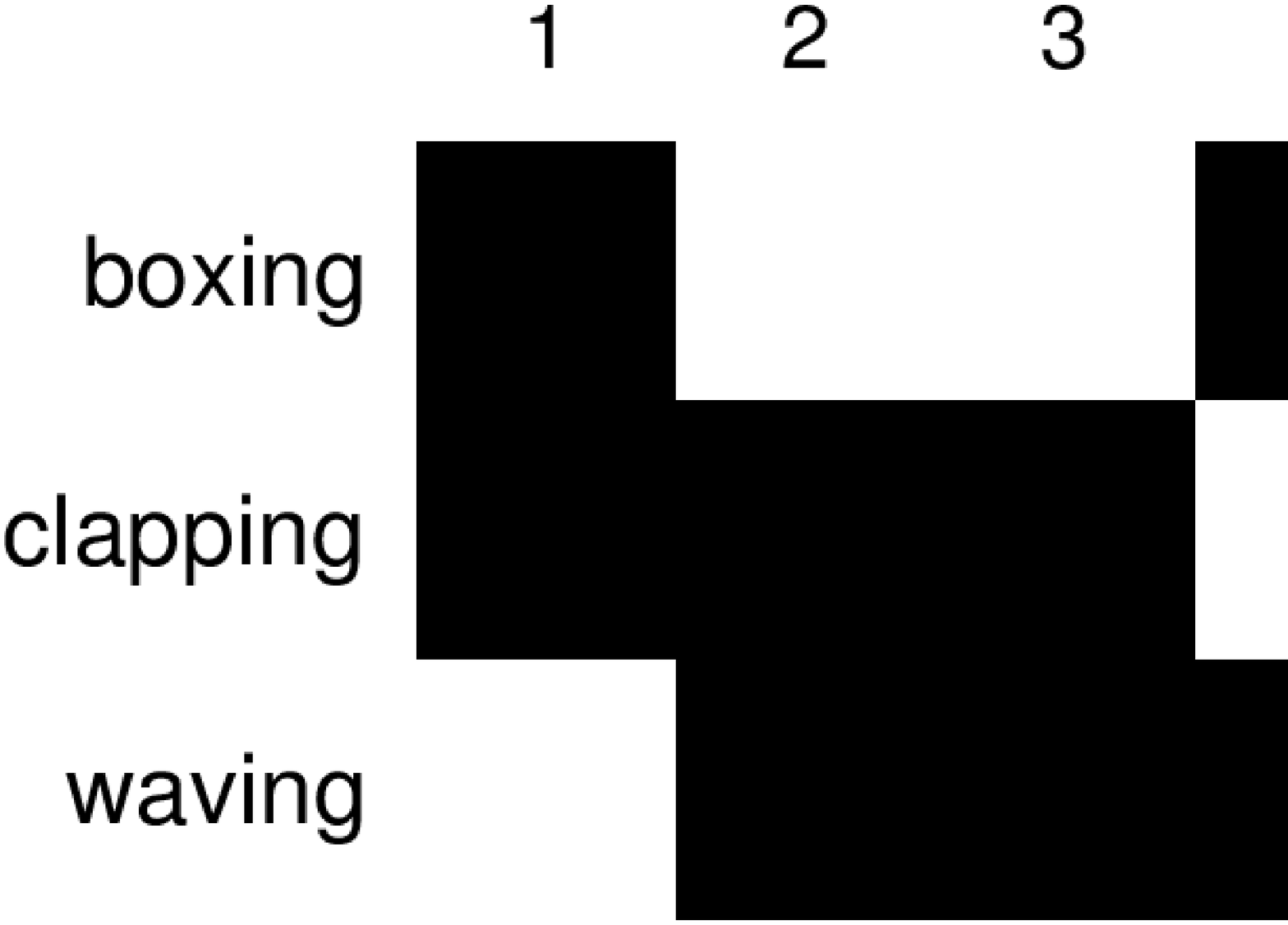}
\label{fig:KTHCM0}}
\subfigure[\small $\boldsymbol\theta^\mathrm{class}$ with SLDA]{
\includegraphics[width=0.33\columnwidth]{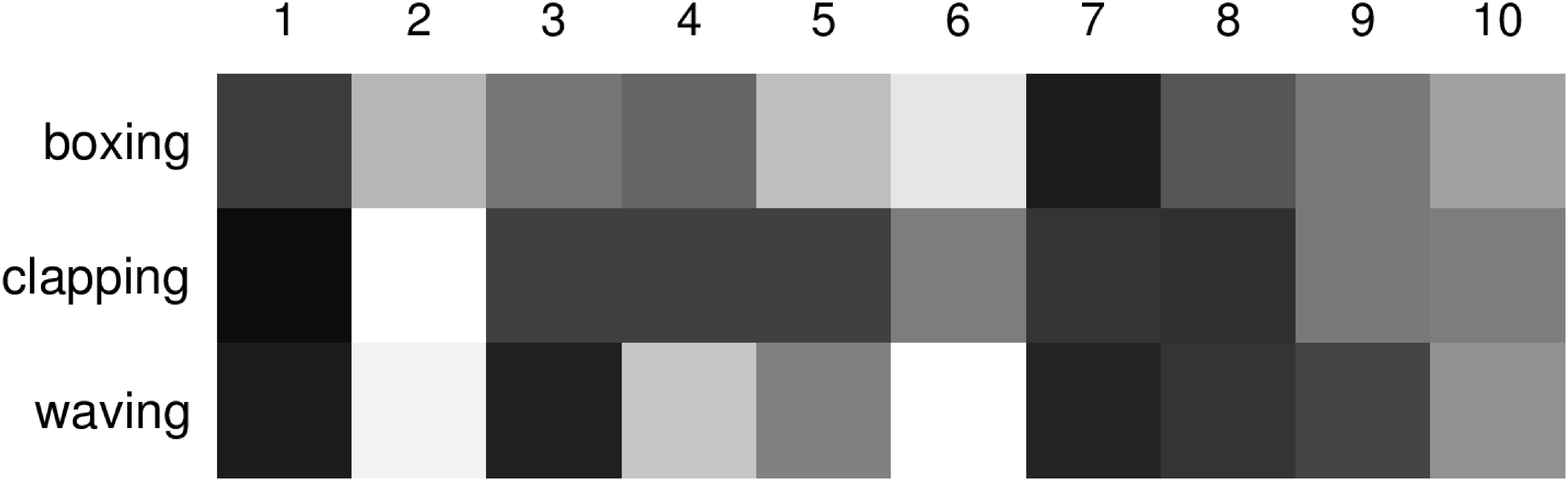}
\label{fig:KTHCMSLDA}}
\subfigure[\small $\boldsymbol\theta^\mathrm{class}$ with factorized LDA]{
\includegraphics[width=0.33\columnwidth]{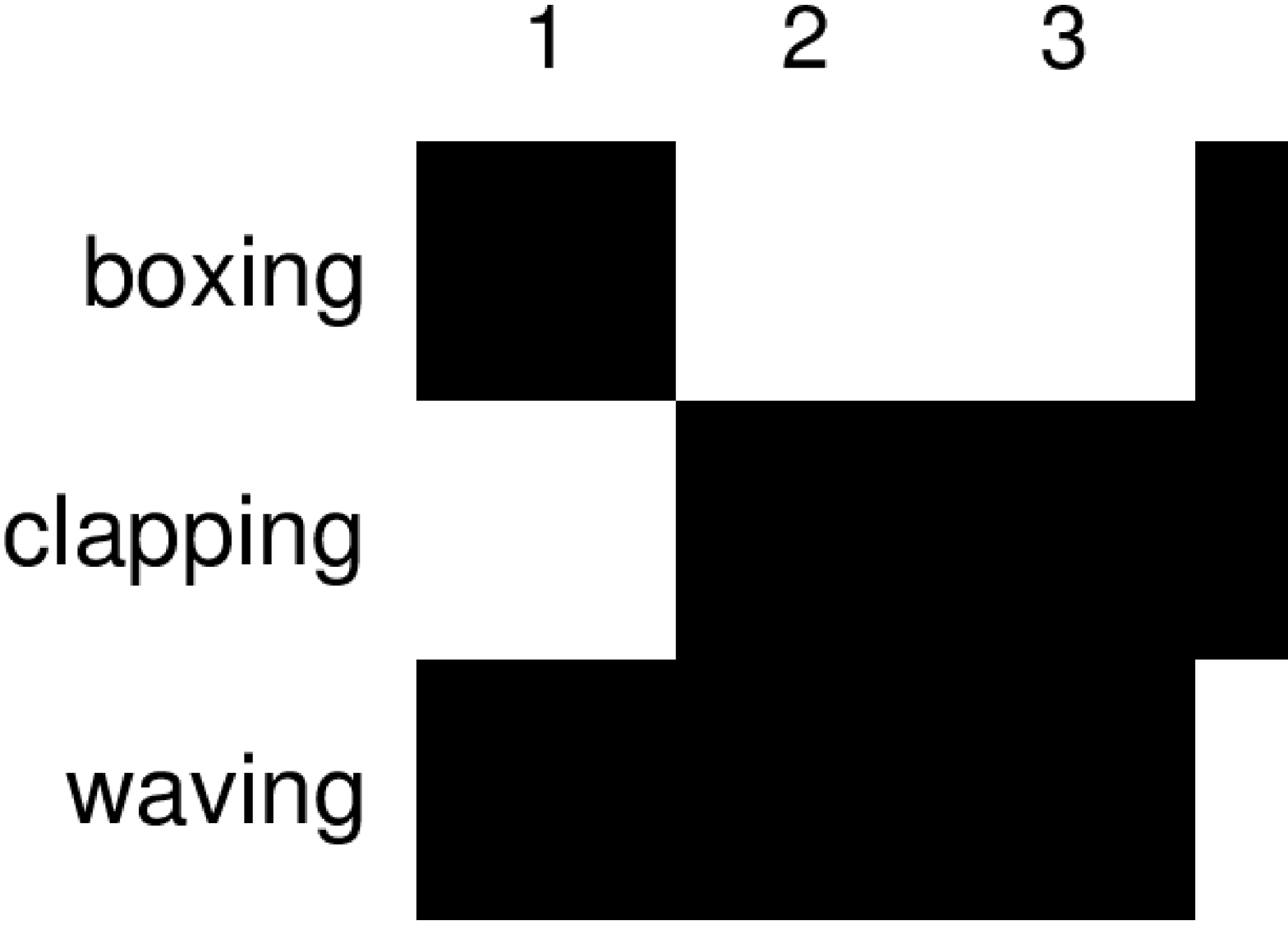}
\label{fig:KTHCM1}}}
\vspace{-3mm}
\caption{ \small Action dataset. (a) Regular LDA topic distribution
marginalized over class $\boldsymbol\theta^\mathrm{class}$, topics sorted in
ascending order of class-specif{}icity. (b) SLDA topic distribution
marginalized over class $\boldsymbol\theta^\mathrm{class}$, topics sorted in
ascending order of class-specif{}icity. (c) Factorized LDA topic distribution
marginalized over class $\boldsymbol\theta^\mathrm{class}$, topics sorted in
ascending order of class-specif{}icity, red line indicating partition between
$\boldsymbol\theta^p$ and $\boldsymbol\theta^s$.}
\label{fig:KTH}
\vspace{-2mm}
\end{figure}

Figure \ref{fig:KTH} shows the topic distributions corresponding to these three models. We can see that Factorized LDA is able to model the class-dependent information (left of the red line) and the class-independent information ( right of the red line), which makes it be able to archive better performance in noisy data. For the regular LDA, although the topics are not shared, however, it models all the information and assigned that to different classes with new topics which made the topics themselves became noisy. So does SLDA which models the "noise" as the useful topics.

Factorized LDA gives an accuracy of 65.22\%, which is far better than both regular LDA, 38\%, and SLDA, 51.33\%. This conf{}irms that the f{}indings of the toy experiment above applies to realistic settings as well. Confusion matrices  are shown in Figures \ref{fig:KTHCM0},  \ref{fig:KTHCMSLDA}, and \ref{fig:KTHCM1} respectively.

%% file: discussion.tex
\section{Conclusions}
\label{sec:discussion}
\vspace{-2mm}

We present a factorized latent topic model,  which explicitly represents aspects
of the data which are not correlated with model state. Specif{}ically, we train an
LDA class model with an additional factorizing prior, which encourages topics to either be
very class-specif{}ic or evenly shared among classes. The topic space $\boldsymbol\theta$ is
thus partitioned into one part $\boldsymbol\theta^p$ whose topics are private to certain
classes, and another part $\boldsymbol\theta^s$ with topics shared between classes. Only 
$\boldsymbol\theta^p$ contributes effectively to classif{}ication. 

Experiments show the factorized LDA model  to give consistently better classif{}ication
performance and sparser topic representations than both a regular LDA model \cite{feifei05} and SLDA \cite{blei10}.
Sparse representations are advantageous for large datasets since they save
storage space and computation time during classif{}ication.

Future work includes investigating the effect of  this factorization prior on
other topic models, such as HDP, and to integrate the prior into models with
multiple data views, such as in \cite{jia11,Virtanen12,cwang09}.